\documentclass{article}

\usepackage[nonatbib, preprint]{neurips_2022}

\usepackage[utf8]{inputenc} % allow utf-8 input
\usepackage[T1]{fontenc}    % use 8-bit T1 fonts
\usepackage{hyperref}       % hyperlinks
\usepackage{url}            % simple URL typesetting
\usepackage{booktabs}       % professional-quality tables
\usepackage{amsfonts}       % blackboard math symbols
\usepackage{nicefrac}       % compact symbols for 1/2, etc.
\usepackage{microtype}      % microtypography
\usepackage{xcolor}         % colors
\usepackage{graphicx}
\usepackage{tabularx}
\usepackage{multicol}
\usepackage[
  backend=biber,
  doi=false]{biblatex}
\title{Multimodal Representation Learning With Text and Images}

\author{
\and
  Aishwarya Jayagopal
  \texttt{e0674492@u.nus.edu}
  \and
  Ankireddy Monica Aiswarya
  \texttt{e0674486@u.nus.edu}
  \and
  Ankita Garg
  \texttt{e0674471@u.nus.edu}
  \and
  Srinivasan Kolumam Nandakumar
  \texttt{e0674494@u.nus.edu}
\and
}

\begin{document}

\maketitle

\begin{abstract}
In recent years, multimodal AI has seen an upward  trend as researchers are integrating data of different types such as text, images, speech into modelling to get the best results. This project leverages multimodal AI and matrix factorization techniques for representation learning, on text and image data simultaneously, thereby employing the widely used techniques of Natural Language Processing (NLP) and Computer Vision. The learnt representations are evaluated using downstream classification and regression tasks. The methodology adopted can be extended beyond the scope of this project as it uses Auto-Encoders for unsupervised representation learning.
\end{abstract}

\section{Introduction}
Supervised learning models are trained using a labelled dataset. And as a result, these models learn features in data while mapping it to the respective output labels. There are several drawbacks of using Supervised learning. If the input to the model is irregular or noisy, it will give drastically inaccurate results. It requires the training set to include all possible examples that belong to a class, else it will output a wrong class. To overcome these limitations,  we adopt unsupervised representation learning techniques to learn abstract representations in raw data. The theory used is to reduce high-dimensional data to lower dimensions while retaining essential information so as to discover new patterns in data. This also reduces the manual task of labelling the dataset. 

Visual data such as images, videos usually require a domain expert to identify and define the certain qualities algorithmically, like the transformations to be used. To make the model invariant to the input, Deep Neural Networks (DNNs) are used to learn representation of input data such as text, images. This is done using an iterative process, wherein DNNs learn to make good predictions by iteratively updating behaviour of neurons in all layers. This requires the domain experts to define the target labels and the accuracy they want from the network. The output of the current layer will serve as inputs to the subsequent layer, and the final-most layer outputs predicted labels. Once the network is performing with the desired accuracy, the iterative process can come to a halt. This network can now be fed as a pre-trained network to a different learning task, as the penultimate layer of this network outputs the abstract representations for new task, also known as transfer learning. 

In domains such as healthcare, input data can come from various modalities, such as X-ray images, clinical records, clinical features etc. A patient can be described by all of these features simultaneously. As a result, to represent an entity like a patient, would need methods that can learn jointly from these varied sources of data, which also vary in modalities. Multi-modal representation learning is thus an area of interest in such domains.

As we often work on large unlabelled datasets, we use unsupervised representation learning on multi-modal data, consisting of images and text in particular, in order to efficiently reduce the problem of manual feature engineering, which is a computationally expensive process. We leverage the capabilities of autoencoders - vanilla and convolutional- along with matrix factorization methods to achieve this.

Matrix factorization methods have been used in learning representations from collections of matrices. However, they require the inputs to be in the form of matrices. In case of inputs like images, additional processing would be necessary to convert such higher dimensional tensors into matrices. In this paper, we propose a novel architecture that can take in a combination of tensors and matrices as inputs, and learn representations for the associated entities jointly, in an end-to-end fashion. 

In the subsequent sections, we will first survey existing matrix factorization methods that work on an arbitrary collection of matrices, the details of our proposed architecture, the details of the dataset we use for evaluating our model, the model development and the comparison between existing methods and our model. We conclude with how this model can be enhanced further. 
% To learn deep network representations of images, Auto-Encoders are being used. Simply put, they encode and decode an image. It does so by reducing the dimension of image, which will serve as the lower-dimensional representation of the image, and then use this representation to reconstruct the high-dimensional original image. The network will be trained to reduce the reconstruction loss to capture the essential parts of the image. 

\section{Literature Survey}
For multi-modal data setup, matrix factorization methods are generally preferred. Some of the common methods found to be useful in learning representations for entities from a collection of matrices are Collective Matrix Factorization\cite{singh2008relational}, Data Fusion by Matrix Factorization\cite{vzitnik2014data} and Deep Collective Matrix Factorization \cite{mariappan2019deep}.  These methods learn entity specific representations for the entities along the rows and columns of the matrices and use them to reconstruct the matrices. However, these methods expect the input to be a collection of matrices, and tensors are not supported. Our model overcomes this limitation by combining the reconstruction capabilities of matrix factorization and the ability of convolutional autoencoders to work on tensor inputs. Thus, with our proposed architecture, we can jointly learn representations for entities based on inputs of any dimensions(matrices or tensors).

CMF factorizes a matrix into two factors – one for the row entity and another for the column entity. If $U^{[r_{m}]}$ represents the row entity representation and $U^{[c_{m}]}$ represents the column entity representation for a matrix $X_{m}$, with $r_{m}$ as the row and $c_{m}$ as the  column entity, $X_{m} = f_{m}(U^{[r_{m}]}.U^{[c_{m}]T})$, where $f_{m}$ is a matrix specific link function. However $X_{m}$ can only take scalar values in each of its cells.

DFMF performs a tri-factorization of a matrix R as $R = G_{i}S_{ij}G_{j}^{T}$ where $S_{ij}$ is a factor describing the relation between the row and column entities. $G_{i}$ and $G_{j}$ are the row and column representations for the relation matrix R. This also takes in as input a collection of matrices.  Also, here R is assumed to be a binary matrix, where each cell denotes the degree of a relationship – presence or absence.

DCMF concatenates the matrices sharing the same entity, and uses autoencoders to factorize the concatenated matrices into entity representations. Here, both real and binary values are possible for the cells making up the matrices, however the collection can only comprise of matrices, due to the concatenation process. For example, if an entity E is described by both a one-dimensional vector and by a 3-dimensional image, concatenating the two views would be difficult.

The model we propose can handle the presence of two differing dimensional views of the same entity, like a book denoted by a textual representation (which can be represented using Word2Vec\cite{mikolov2013efficient}) and by images. This finds extensive application in many domains like the healthcare sector, where patients can be described by both their clinical records as well as their X-ray records.

\section{Proposed Architecture}
\label{Proposed_Architecture}
\begin{figure}
    \centering
    \includegraphics[width=0.9\linewidth]{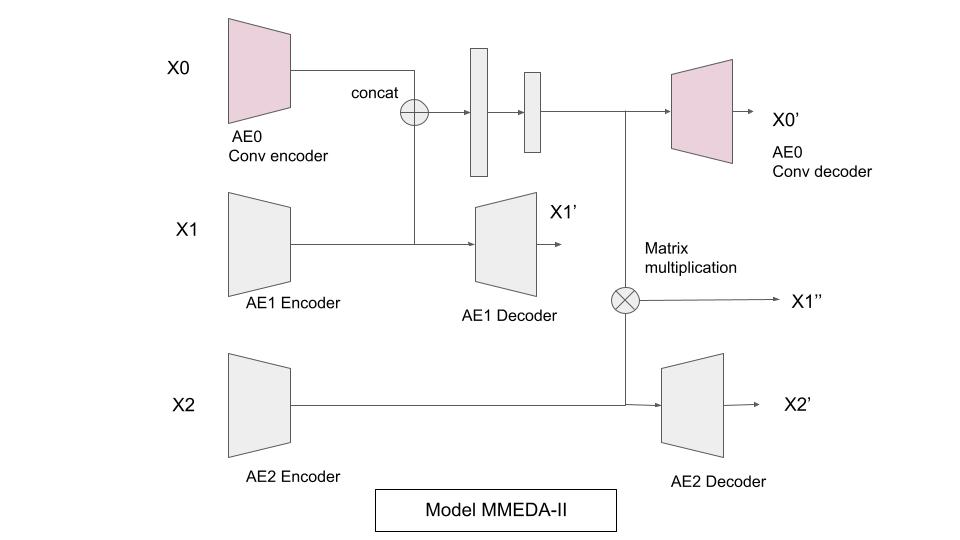}
    \caption{MMEDA-II architecture}
    \label{fig:mmeda_2}
\end{figure}

We propose an end-to-end architecture for multi-modal representation learning. Figure \ref{fig:mmeda_2} shows the proposed architecture - Multi-Modal Encoder Decoder Architecture - II(MMEDA-II). To explain our architecture, we will be considering the case of two inputs - a matrix and a three-dimensional tensor.

The model accepts three input matrices: X0 (of dimensions m x n1 x k), X1 (of dimensions m x n2). A third matrix X2 ( of dimensions n2 x m) is created by transposing X1. In this case, the $i^{th}$ row in X0 and X1 correspond to the attributes of the same entity. X0 is passed through a convolutional autoencoder (AE0) and X1 and X2 are passed through a feed forward vanilla autoencoder (AE1 and AE2 respectively). The decoding output from autoencoders AE0, AE1 and AE2 are represented by X0’, X1’ and X2’ respectively.

Because we want the final representation for a data point to include information from both matrices X0 and X1, output from the encoder layer of AE0 is concatenated with output from the encoder layer of AE1 and passed through a feed forward network to obtain the final representation. This final representation is multiplied with the transpose of the output of the encoder layer of AE2 to get matrix X1’’.

\subsection{Model Training}
The training data is represented as matrices M0 and M1, where M0 represents multi-dimensional attributes of the data points (e.g., image data) and M1 represents one-dimensional attributes of the data points (e.g., text data vectors). The data points are represented by each row of the matrices, and the attributes are represented by each column.

The model is trained in batches. Each batch consist of b data points. For each batch i, model input $X0 = M0[(i *b) : (i*b) + b]\ and\ X1 = M1[(i *b) : (i*b) + b]$ and X2 is computed by iterating over the transpose of M1. In each iteration j, $X2 = M2[(j*b) : (j*b)+b]$ where M2 is the transpose of M1. 

After each iteration j, the model parameters are learned by minimizing reconstruction loss between model input and output. The loss function is given by equation\ref{mmeda2_loss_equation}.

\begin{equation}
    L_{total} = L_{MSE}(X0', X0) + L_{MSE}(X1', X1) + L_{MSE}(X2', X2)+L_{MSE}(X1'', X1) \\
\label{mmeda2_loss_equation}
\end{equation}

\subsection{Getting representations}
To get representations for the data points, the multi-dimensional attributes are passed through AE0 encoder and the one-dimensional attributes are passed through AE1 encoder. The output from both encoders is concatenated and passed through the feed forward network. The output of the feed forward network gives the final representation for the corresponding data point. 

\section{Dataset Description}
The dataset  used in this project is the Goodreads dataset from \cite{WanM18} and \cite{WanMNM19}, with about 2.3 million books found on the Goodreads website, with cover images and features like average ratings, reviews, genre categories and book descriptions as some of the available features.  

In this dataset, we are interested in learning lower dimensional representations of books, based on the book descriptions and the cover images.

\subsection{Preprocessing}
Books that had a description present, average rating greater than zero and with language code set as English were retained. We considered a sample of 5000 books from the resulting dataset and used this for learning representations. The genre of the 5000 records were converted into a binary label - fiction and non fiction. The genre for a book was considered fiction if the original genre was one of fiction, romance, fantasy, paranormal and young-adult, and non fiction otherwise. Overall, the dataset has 3188 books under the fiction genre and 1812 non fiction books. 

\subsection{Word2Vec}
The description of books in text format needed to be converted into vectors, that could be used for training neural nets. Two commonly used word embeddings are Glove\cite{pennington2014glove} and Word2vec\cite{mikolov2013efficient}. Word2Vec is chosen in this project and the embedding model is trained on the text descriptions. For each description, the word embeddings of all the words are taken and averaged to get the word embedding of that particular description.

\subsection{Pre-trained convolutional neural network}
For a comparative study, in this project a set of pre-trained convolutional neural networks are used to acquire representations of the cover images of books. The following are the pre-trained models considered:

\begin{itemize}
    \item Google Net \cite{szegedy2015going}
    \item Inception Net \cite{szegedy2016rethinking}
    \item ResNet \cite{he2016deep}
    \item VGG \cite{simonyan2014very}
\end{itemize}

The final classification layer is removed in all the pretrained models and all the images are passed through the models, to obtain a 1000-dimensional representation for each image.

\section{Experiment Settings}
To conduct our experiment, the available dataset was divided into 80-20 train and test splits, with 4000 books in the train and 1000 books in the test splits. The complete set of 5000 books, with their Word2Vec representations and cover images were passed through the models described above to obtain lower dimensional representations, of 50 dimensions. The learnt representations for the 4000 books in the train set were used to train the downstream task, and the representations for the 1000 test books were used to evaluate the representations.

The downstream tasks were the prediction of the book average rating, modelled as a regression task, and the genre classification, modelled as a classification task. The regression was done using the Linear regression package in scikit-learn and classification using RandomForest from scikit-learn. A maximum depth 2 and random state 0 were used for the RandomForest classifier. We consider the F1 score, accuracy, precision and recall as the evaluation metrics for classification and mean square error for regression.

\section{Model Development}
Having obtained the processed dataset, we first used a supervised multi layer perceptron (subsection \ref{sec:MLP}) as a baseline to compare our models against. We started with a model that used matrix factorization (subsection \ref{sec:pretrained_model}), similar to the algorithm in \cite{mariappan2019deep}, on the pre-trained one-dimensional image representations and Word2Vec representations. Since these pre-trained models are trained to perform a specific task, and our goal was to obtain image representations that are not mapped to any label, we replaced the pre-trained models with convolutional autoencoders. This lead to the model described in subsection \ref{sec:convAE}. In the subsequent models, we modify the architecture to train both the convolutional autoencoders and the matrix factorization architecture in an end-to-end fashion (subsections \ref{sec:mmeda_m1} and \ref{sec:mmeda_m2}). 

\begin{figure}
    \centering
    \includegraphics[width=0.8\textwidth]{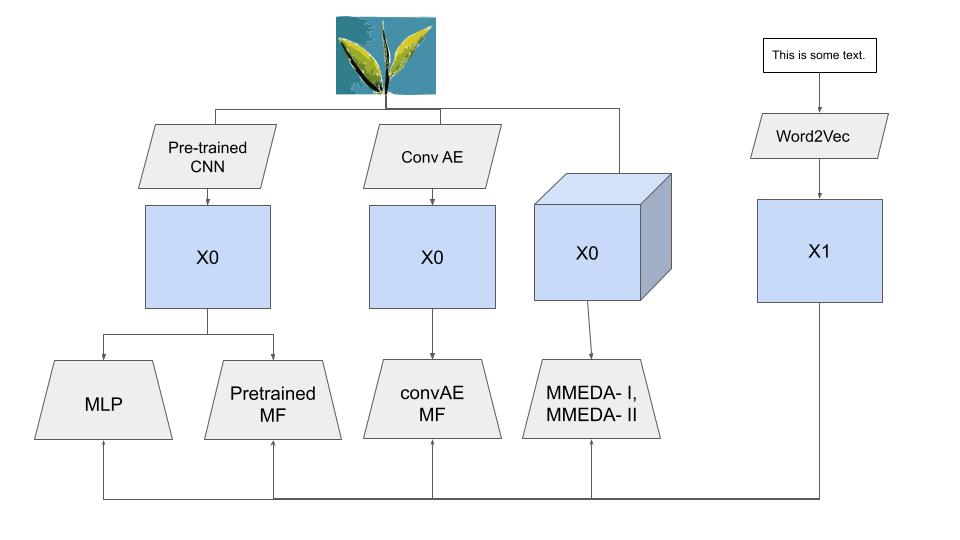}
    \caption{Overall pipeline followed during model development}
    \label{fig:flowchart}
\end{figure}

The figure \ref{fig:flowchart} shows the overall pipeline followed.

\subsection{Multi Layer Perceptron}
\label{sec:MLP}
The input to the MLP model is obtained by concatenating the Word2Vec tensors representing the description for each book and the image representations generated. The image representations varied depending on the pre-trained model used so as to compare which network was giving the best results for the dataset used. 

For the task of classification of genres, Adam optimiser with a learning rate of 5e-4 was chosen after hyperparameter tuning. Cross Entropy Loss was used as it is widely used for classification tasks. For the task of average rating prediction using regression, Stochastic Gradient Descent optimiser with a learning rate of 1e-4 was found to perform the best during hyperparameter tuning. Mean Square Error loss was used as the loss function. MLP models serve as the baseline for further evaluating other architectures that were implemented.

\subsection{Pre-trained CNN Matrix Factorization}
\label{sec:pretrained_model}
Using the image representations of the book covers along with the Word2Vec representations, an encoder-decoder based matrix factorization algorithm was trained to learn the book representations jointly from the two input matrices. The embedding size was set to 50 and an encoder was used to reduce the concatenated input(image representation from pre-trained networks + Word2Vec embedding) to the embedding size. A decoder resets the embedded layer to the size of the input. The output of the encoder is the row entity representation. To obtain the column entity representation for both input matrices, their column transpose was passed through two distinct autoencoders. The resulting row and column representations for each matrix is multiplied to reconstruct the original input matrices.  Mean square loss is calculated between the output and the input, to train the network. A batch size of 50 and a learning rate of 1e-4 was set with a convergence criteria of 1e-4.  The training converged after 160 epochs.

\subsection{Convolutional Autoencoder Based Matrix Factorization}
\label{sec:convAE}
A convolutional auto encoder was trained to learn the representation of the book cover images. The images are passed through two convolutional layers with max pooling and the output from the second layer is flattened and passed through 2 linear layers to get an embedding size of 200. The decoder reconstructs the image and the final mean square loss is calculated to train the network. A batch size of 32 was used and network was trained for 100 epochs. The resulting image representation is used instead of the pre-trained model image representation, and the same procedure is followed as described in Section \ref{sec:pretrained_model}.

\begin{figure}
    \centering
    \includegraphics[width=0.8\textwidth]{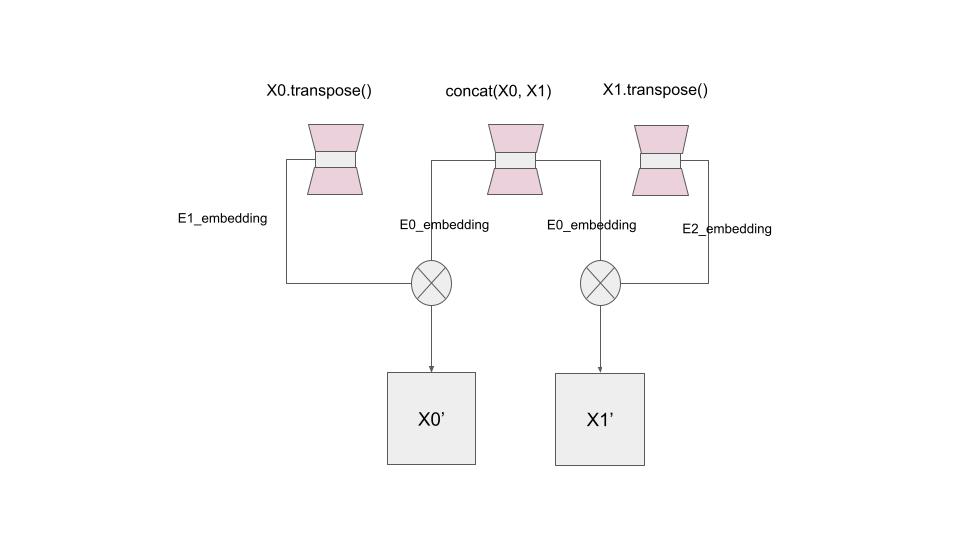}
    \caption{Matrix factorization architecture using autoencoders. The inputs here are assumed to be one-dimensional vectors for each entity.}
    \label{fig:matrix_factorization}
\end{figure}

Both methods, using convolutional autoencoder and pre-trained CNN for obtaining image representations, use the architecture in figure \ref{fig:matrix_factorization} for matrix factorization.

\subsection{Multi-Modal Encoder Decoder Architecture-I (MMEDA-I)}
\label{sec:mmeda_m1}
\begin{figure}
    \centering
    \includegraphics[width=0.9\linewidth]{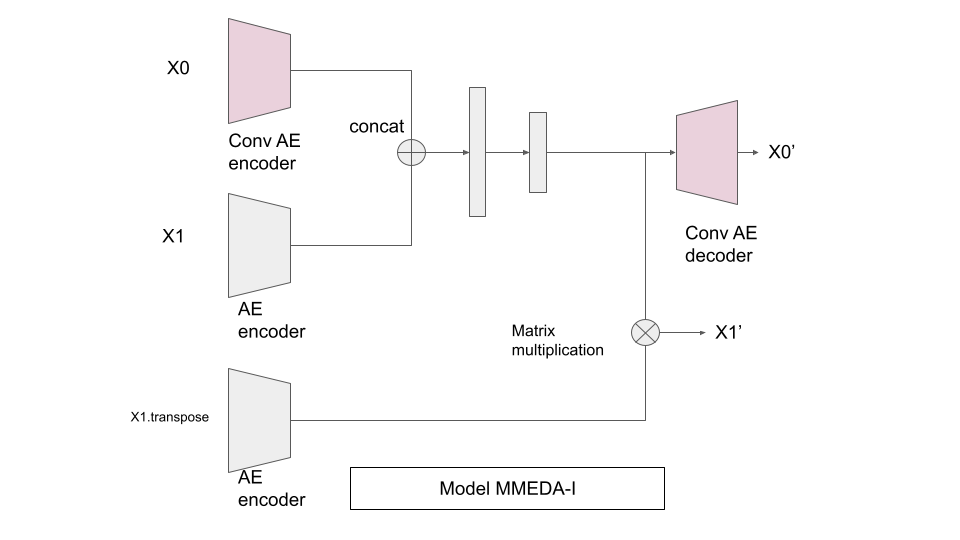}
    \caption{MMEDA-I architecture, for a set of two matrices, X0 made up of images and X1 of text representations. Convolutional and vanilla encoder-decoders are used to learn representations.}
    \label{fig:mmeda_1}
\end{figure}
In this phase of our model development, we aimed to use the one-dimensional representations(Word2Vec for text inputs) and the tensor view (images) jointly, in an end-to-end fashion. To this end, we used vanilla autoencoders and convolutional autoencoders for each input type respectively. The architecture in Figure \ref{fig:mmeda_1} shows the setup for two inputs – a matrix and a tensor where the $0^{th}$ dimension holds the entity of our interest. 

The matrix is passed through a vanilla encoder(to get row entity representations), its transpose through another vanilla encoder(to get column entity representations), while the tensor is passed through a convolutional encoder. Since the representation for the entity of interest must be jointly learnt from both the inputs(matrix and tensor), the lower dimensional representations from both encoder feed forward networks are concatenated together and passed through another feed forward network, to obtain a final lower dimensional representation of the entity. This lower dimensional representation is used to reconstruct the tensor by passing it through a convolutional decoder and is used to reconstruct the matrix by multiplying with the column entity representation, obtained from the column encoder. The combined loss minimised is the sum of reconstruction losses(MSE loss) for both the tensor and the matrix, as shown in the equation below. This method can be extended to an arbitrary collection of tensors and matrices.
\begin{equation}
    L_{total} = L_{MSE}(X0', X0) + L_{MSE}(X1', X1)
\end{equation}

\subsection{MMEDA-II}
\label{sec:mmeda_m2}
MMEDA-II is an extension of model MMEDA-I (subsection \ref{sec:mmeda_m1}). The details for the architecture can be found in section \ref{Proposed_Architecture}.

\section{Results}
From our experiments, we evaluated the results of the proposed model and the models trained during the model development process, against popular matrix factorization methods like CMF. For CMF to work, we obtain input matrices by passing the image tensor object through pre-trained CNN networks like GoogleNet, InceptionNet, ResNet and VGG. We also compare the methods against a baseline multi-layer perceptron. The results are shown in Table \ref{tab:results}.  The representations learnt from the various architectures are used to perform a downstream classification and regression task. The F1 score, precision, recall are compared for classification, while the mean square error is used to compare the regression results. The results for the 4 pre-trained CNN models appear quite similar, probably because they are pre-trained on the same dataset, and towards the last layers, the representations learnt could be similar.

\begin{table}
\centering
\begin{tabular}[h]{ c  c c  c c  c c }
\toprule
Model & \multicolumn{2}{c}{MLP} & \multicolumn{2}{c}{Pre-trained MF} & \multicolumn{2}{c}{CMF} \\
\midrule
  &  F1  &  MSE  &  F1  &  MSE  &  F1  &  MSE  \\
\midrule
Google Net & 0.7748 & 0.2424 & \textbf{0.7782} & 0.1778 & \textbf{0.7782} & \textbf{0.1745} \\
\midrule
Inception Net & 0.7352 & 0.2991 & \textbf{0.7782} & 0.1763 & \textbf{0.7782} & \textbf{0.1745} \\
\midrule
VGG & 0.7459 & 0.2943 & \textbf{0.7782} & \textbf{0.1763} & \textbf{0.7782} & 0.1764 \\
\midrule
ResNet & 0.7503 & 0.1789 & 0.7772 & 0.1783 & \textbf{0.7853} & \textbf{0.1768} \\
\bottomrule\\
\end{tabular}
\caption{Results obtained from multi-layer perceptrons(MLP), pre-trained CNN matrix factorization(pretrained MF) and Collective Matrix Factorization(CMF), in downstream classification and regression tasks. F1 score is used to evaluate classification performance and MSE for regression.}
\label{tab:results}
\end{table}

\begin{table}
\centering
    \begin{tabular}{ c  c c c c  c }
    \toprule
    Architecture & \multicolumn{4}{c}{Classification} & \multicolumn{1}{c}{Regression} \\
    \midrule
     & F1 & Accuracy & Precision & Recall & MSE \\
    \midrule
    CMF ResNet & \textbf{0.7853} & 0.655 & 0.6505 & 0.99 & 0.1768 \\
    \midrule
    ConvAE MF & 0.7782 & 0.637 & 0.637 & \textbf{1} & 0.1757 \\
    \midrule
    MMEDA-I & 0.7639 & 0.686 & 0.733 & 0.7974 & 0.1753 \\
    \midrule
    MMEDA-II & 0.7727 & \textbf{0.70} & \textbf{0.7467} & 0.8006 & \textbf{0.1737} \\
    \bottomrule\\
    \end{tabular}
    \caption{Results obtained for downstream classification and regression, for best performing CMF model and the proposed methods obtained during model development.}
    \label{tab:results_part2}
\end{table}

From the methods in Table \ref{tab:results}, we obtain the best performing method(CMF using ResNet image representations) and compare it against the proposed method, along with the intermediate methods obtained during model development. These results are shown in Table \ref{tab:results_part2}. Although unsupervised, the representations learnt from our proposed architecture appears to do well. The joint learning from the different modalities, directly from the input tensors, could be one reason for this. Being an unsupervised learning method, the learnt representations are reusable and can be used for any downstream task, including clustering. The availability of fewer data points could be one explanation for the relatively poor performance of MLP network, though it is a supervised learning framework. Thus, it can be concluded that the  proposed method can do well while learning representations from a collection of matrices and tensors. 

\section{Conclusions}
Thus, in conclusion, we see that our proposed method can learn representations jointly from a collection of matrices and tensors. This is especially useful in domains such as healthcare, where a patient can be viewed from different modalities, like X-rays, clinical records etc. The proposed architecture is found to outperform benchmark matrix factorization methods and supervised multi-layer perceptrons.

Representing the text inputs using transformer models, instead of averaging Word2Vec representations, which can be trained along with the representation learning, and extending this architecture to work on sequential data, like videos, are other directions worth exploring. Replacing vanilla autoencoders with convolutional autoencoders in existing matrix factorization architectures and using variational autoencoders instead of vanilla autoencoders can also be explored.

\printbibliography

@inproceedings{singh2008relational,
  title={Relational learning via collective matrix factorization},
  author={Singh, Ajit P and Gordon, Geoffrey J},
  booktitle={Proceedings of the 14th ACM SIGKDD international conference on Knowledge discovery and data mining},
  pages={650--658},
  year={2008}
}

@article{vzitnik2014data,
  title={Data fusion by matrix factorization},
  author={{\v{Z}}itnik, Marinka and Zupan, Bla{\v{z}}},
  journal={IEEE transactions on pattern analysis and machine intelligence},
  volume={37},
  number={1},
  pages={41--53},
  year={2014},
  publisher={IEEE}
}

@article{mikolov2013efficient,
  title={Efficient estimation of word representations in vector space},
  author={Mikolov, Tomas and Chen, Kai and Corrado, Greg and Dean, Jeffrey},
  journal={arXiv preprint arXiv:1301.3781},
  year={2013}
}

@inproceedings{szegedy2015going,
  title={Going deeper with convolutions},
  author={Szegedy, Christian and Liu, Wei and Jia, Yangqing and Sermanet, Pierre and Reed, Scott and Anguelov, Dragomir and Erhan, Dumitru and Vanhoucke, Vincent and Rabinovich, Andrew},
  booktitle={Proceedings of the IEEE conference on computer vision and pattern recognition},
  pages={1--9},
  year={2015}
}

@inproceedings{szegedy2016rethinking,
  title={Rethinking the inception architecture for computer vision},
  author={Szegedy, Christian and Vanhoucke, Vincent and Ioffe, Sergey and Shlens, Jon and Wojna, Zbigniew},
  booktitle={Proceedings of the IEEE conference on computer vision and pattern recognition},
  pages={2818--2826},
  year={2016}
}

@inproceedings{he2016deep,
  title={Deep residual learning for image recognition},
  author={He, Kaiming and Zhang, Xiangyu and Ren, Shaoqing and Sun, Jian},
  booktitle={Proceedings of the IEEE conference on computer vision and pattern recognition},
  pages={770--778},
  year={2016}
}

@article{simonyan2014very,
  title={Very deep convolutional networks for large-scale image recognition},
  author={Simonyan, Karen and Zisserman, Andrew},
  journal={arXiv preprint arXiv:1409.1556},
  year={2014}
}

@article{mariappan2019deep,
  title={Deep collective matrix factorization for augmented multi-view learning},
  author={Mariappan, Ragunathan and Rajan, Vaibhav},
  journal={Machine Learning},
  volume={108},
  number={8},
  pages={1395--1420},
  year={2019},
  publisher={Springer}
}

@inproceedings{WanM18,
  author    = {Mengting Wan and
               Julian J. McAuley},
  editor    = {Sole Pera and
               Michael D. Ekstrand and
               Xavier Amatriain and
               John O'Donovan},
  title     = {Item recommendation on monotonic behavior chains},
  booktitle = {Proceedings of the 12th {ACM} Conference on Recommender Systems, RecSys
               2018, Vancouver, BC, Canada, October 2-7, 2018},
  pages     = {86--94},
  publisher = {{ACM}},
  year      = {2018},
  url       = {https://doi.org/10.1145/3240323.3240369},
  doi       = {10.1145/3240323.3240369},
  bibsource = {dblp computer science bibliography, https://dblp.org}
}

@inproceedings{WanMNM19,
  author    = {Mengting Wan and
               Rishabh Misra and
               Ndapa Nakashole and
               Julian J. McAuley},
  editor    = {Anna Korhonen and
               David R. Traum and
               Llu{\'{i}}s M{\`{a}}rquez},
  title     = {Fine-Grained Spoiler Detection from Large-Scale Review Corpora},
  booktitle = {Proceedings of the 57th Conference of the Association for Computational
               Linguistics, {ACL} 2019, Florence, Italy, July 28- August 2, 2019,
               Volume 1: Long Papers},
  pages     = {2605--2610},
  publisher = {Association for Computational Linguistics},
  year      = {2019},
  url       = {https://doi.org/10.18653/v1/p19-1248},
  doi       = {10.18653/v1/p19-1248},
  bibsource = {dblp computer science bibliography, https://dblp.org}
}

@inproceedings{pennington2014glove,
  title={Glove: Global vectors for word representation},
  author={Pennington, Jeffrey and Socher, Richard and Manning, Christopher D},
  booktitle={Proceedings of the 2014 conference on empirical methods in natural language processing (EMNLP)},
  pages={1532--1543},
  year={2014}
}
\appendix

\section{Appendix}
The code implementing the architecture described in this paper is available at \url{https://github.com/Srini-98/CS5260-Neural-Networks-2}.

\end{document}